\documentclass{article}
\usepackage{ijcai25}
\usepackage{times}
\usepackage{array}
\usepackage{diagbox}
\usepackage{soul}
\usepackage{bm}
\usepackage{amsfonts}
\usepackage{url}
\usepackage[hidelinks]{hyperref}
\usepackage[utf8]{inputenc}
\usepackage[small]{caption}
\usepackage{graphicx}
\usepackage{amsmath}
\usepackage{amssymb}
\usepackage{amsthm}
\usepackage{booktabs}

\usepackage{algorithm}
\usepackage{multicol}
\usepackage{rotating}
\usepackage{multirow}
\usepackage{algorithmic}
\usepackage[switch]{lineno}
\usepackage{color,xcolor}

\usepackage{subfigure}

\newcommand{\modelname}{GOOD-MIA}
\usepackage[skip=2pt]{caption}
\definecolor{mygray}{RGB}{230,230,230}

\title{An Out-Of-Distribution Membership Inference Attack Approach for Cross-Domain Graph Attacks}

\author{
Jinyan Wang$^{1,2}$
,
Liu Yang\textsuperscript{\rm 1,2},
Yuecen Wei\textsuperscript{\rm 3,4*},
Jiaxuan Si\textsuperscript{\rm 1,2},
Chenhao Guo\textsuperscript{\rm 1,2},\\
Qingyun Sun\textsuperscript{\rm 4},
Xianxian Li\textsuperscript{\rm 1,2},
Xingcheng Fu\textsuperscript{\rm 1,2*}
\affiliations
\textsuperscript{\rm 1}Key Lab of Education Blockchain and Intelligent Technology, Ministry of Education, \\Guangxi Normal University, China\\
\textsuperscript{\rm 2}Guangxi Key Lab of Multi-Source Information Mining and Security, \\Guangxi Normal University, Guilin, China\\
\textsuperscript{\rm 3}School of Software, Beihang University, Beijing, China\\
\textsuperscript{\rm 4}SKLCCSE, School of Computer Science and Engineering, Beihang University, China\\
\emails
\{wangjy612, ylzyg, sijiaxuan03, guochenhao03, fuxc, lixx\}@gxnu.edu.cn, \\
\{weiyc, sunqy\}@buaa.edu.cn
}

\begin{document}

\maketitle

\begin{abstract}
Graph Neural Network-based methods face privacy leakage risks due to the introduction of topological structures about the targets, which allows attackers to bypass the target's prior knowledge of the sensitive attributes and realize membership inference attacks (MIA) by observing and analyzing the topology distribution. 
As privacy concerns grow, the assumption of MIA, which presumes that attackers can obtain an auxiliary dataset with the same distribution, is increasingly deviating from reality. 
In this paper, 
we categorize the distribution diversity issue in real-world MIA scenarios as an Out-Of-Distribution (OOD) problem, 
and propose a novel \textbf{G}raph \textbf{OOD} \textbf{M}embership \textbf{I}nference \textbf{A}ttack (\modelname) to achieve cross-domain graph attacks. 
Specifically, we construct shadow subgraphs with distributions from different domains to model the diversity of real-world data. 
We then explore the stable node representations that remain unchanged under external influences and consider eliminating redundant information from confounding environments and extracting task-relevant key information to more clearly distinguish between the characteristics of training data and unseen data. 
This OOD-based design makes cross-domain graph attacks possible. 
Finally, we perform risk extrapolation to optimize the attack's domain adaptability during attack inference to generalize the attack to other domains. 
Experimental results demonstrate that {\modelname} achieves superior attack performance in datasets designed for multiple domains.

\end{abstract}

\section{Introduction}

Graph Neural Networks (GNNs)~\cite{wu2019simplifying,fu2023hyperbolic} have been widely applied in various practical and potentially high-risk scenarios, such as social networks~\cite{sharma2024survey}, bioinformatics networks~\cite{zhang2021graph}, and medical diagnosis~\cite{boll2024graph}. 
Existing researches~\cite{wu2020comprehensive,velivckovic2017graph} leverage the ability of GNNs to capture structural information and node features to address diverse downstream tasks~\cite{tu2021deep}. 
However, the in-depth mining of data and the powerful representation capabilities of the model also raise serious privacy concerns~\cite{zhang2025disentangled,zhang2024bayesian}. 
\begin{figure}[t]
\centering
\includegraphics[width=0.48\textwidth]{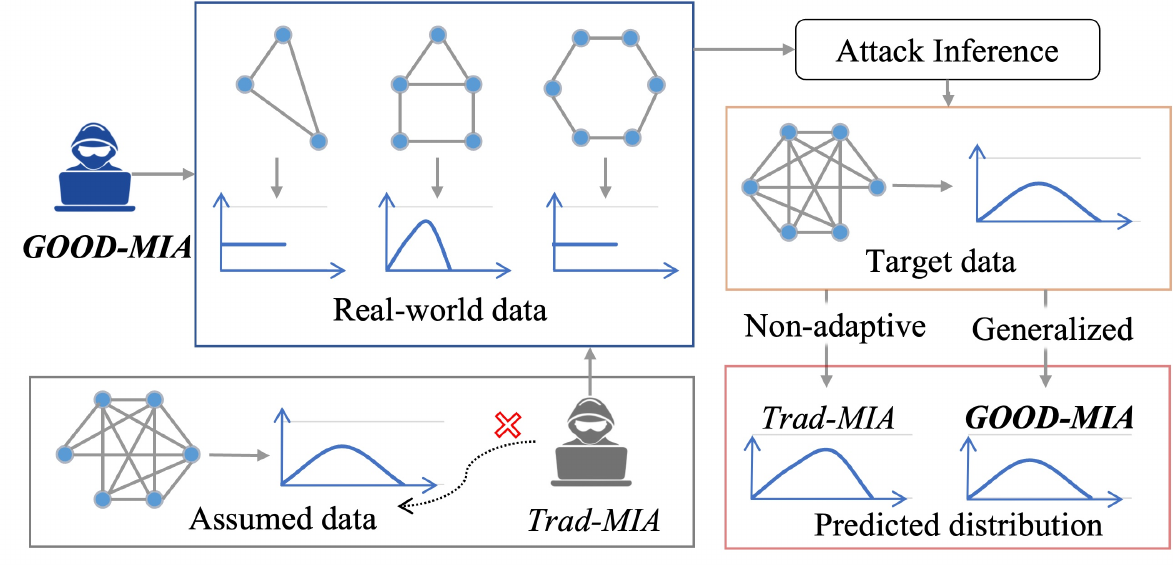}
\caption{
Traditional MIA (Trad-MIA) vs {\modelname}.
}
\label{fig:motiv}
\end{figure}

With the growing concern for personal privacy security~\cite{zhang2024survey,wei2024poincare,li2025rethinking}, graph-structured data has been proven to be highly susceptible to Membership Inference Attacks (MIA) due to its rich associative semantics. 
In MIA, attackers attempt to infer whether a specific node belongs to the training set of the target GNN model, which can lead to severe privacy leakage, especially when the model is trained on sensitive domain datasets. 
For example, when medical diagnosis data is modeled as a graph for training, MIA allows attackers to obtain an individual's health information without having specific details. 
However, the success of traditional MIA~\cite{wei2025prompt} is usually based on the assumption that attackers can access shadow datasets with the same distribution as the target. 
In reality, as shown in Fig.~\ref{fig:motiv}, it is often difficult to obtain similarly distributed data, and there can be biases in data distribution across different environments~\cite{liu2021towards}. These issues lead to diminished attack efficacy due to significant gaps in training data distribution. 
Therefore, cross-domain graph attacks are more practical as they better reflect scenarios where data access is restricted and the data comes from diverse distributions across different domains.

Out-of-distribution (OOD)~\cite{liu2021towards} methods show excellent capabilities in domain adaptation tasks. Existing approaches~\cite{arjovsky2019invariant,krueger2021out} make the model's learned representations consistent across different data distributions through invariant learning, thus providing robustness and generalization to unseen distributions. 
However, graph-structured data may not exhibit high connectivity but a scale-free power-law distribution~\cite{wu2020graph}.
It indicates that the structural differences among nodes and edges in different domains may be significant, requiring the shadow model to account for graph data with substantial distributional differences while mimicking the target model's decisions. 
Therefore, arbitrarily changing the training domain or roughly incorporating graph OOD~\cite{liu2023flood,wu2022handling} methods may lead to a shadow model that fails to adapt to multiple domains, resulting in posterior distribution shifts and affecting the effectiveness of the attack. 

Therefore, to explore the privacy risks faced by models when dealing with data from different distributions, an intuitive idea is to reveal the privacy vulnerabilities of GNNs in OOD inference scenarios by studying the cross-domain graph attack. 
Overall, our research faces the following two challenges: 
Based on the MIA assumptions, existing methods overly rely on specific data features and lack adaptability to the topology of graph OOD data. This results in an inability to distinguish the intrinsic characteristics of training data when faced with confounded distributions, leading to suboptimal attack performance. 
Therefore, \textbf{the key issue lies in simultaneously capturing the common representations of graph features and graph structures across domains, and extending the attack.}
Due to the distribution discrepancies between the shadow and target datasets, existing attack models tend to overfit specific features of the shadow dataset, failing to adequately learn the attributes and structural information directly relevant to downstream tasks. \textbf{This necessitates joint invariant learning to reinforce the acquisition of critical topological information in graph structures. }

To address the above problem, we propose a novel out-of-distribution MIA approach for cross-domain graph attacks, named {\modelname}. 
Specifically, to acquire knowledge from multiple domains during the training, we generate multiple graphs in different environments using an augmentation method~\cite{fu2024hyperbolic}. 
Then, we extract invariant features of the data in multi-domain generalization training to depict the distribution of the training data. 
Moreover, in the model's inference, we constrain sufficient and critical beneficial information for downstream classification tasks and further reinforce invariant representations to maintain the mimicry of the target model's behavior. 
For the attack model, we encourage the equalization of training risks to minimize the likelihood of risk changes when the distribution shifts.
Finally, extensive experiments validate the effectiveness of cross-domain attacks. 
Our contributions are summarized as follows:
\begin{itemize}
    \item To the best of our knowledge, this is the first work to conduct cross-domain attacks against GNNs. It breaks the conventional settings of MIA and reveals the privacy leakage risks of graph models on unknown distributions. 
    \item We propose a novel \textbf{G}raph \textbf{O}ut-\textbf{O}f-\textbf{D}istribution \textbf{M}embership \textbf{I}nference \textbf{A}ttack (\modelname) to achieve cross-domain graph attacks. By capturing invariant representations of cross-domain graph data and constraining the training direction of the model, we mitigate distribution shifts during the risk extrapolation process. 
    \item Comprehensive experiments on multiple real-world datasets demonstrate that {\modelname} leads in cross-domain graph attack performance.
\end{itemize}

\section{Related Work}
\subsection{Membership Inference Attacks on GNN}
MIAs aim to infer whether or not a data sample was used to train a target model~\cite{shokri2017membership}.
Later works~\cite{hayes2017logan,song2019auditing,he2020segmentations} further investigate the feasibility of MIAs in other types of models, such as image generative and segmentation models.
\cite{olatunji2021membership}~first migrated membership inference attack to the graph data, using a shadow training technique. The proposed scheme is based on node-level tasks performed on graph data.
\cite{he2021node}~proposed a scheme for the membership inference attack using the 0-hop subgraph and the 2-hop subgraph, which combined the membership inference attack with the structure of the graph.
However, the adversary requires all needs a shadow dataset that is identically distributed with the target dataset as the auxiliary dataset in the above membership inference attacks.
In practical application scenarios, due to the diversity of data, it is almost impossible to obtain identically distributed data as the auxiliary dataset.
Therefore, the effectiveness of the aforementioned MIAs is likely overestimated~\cite{hintersdorf2021trust}.

\subsection{Out-Of-Distribution Generalization on Graphs}
Previous work~\cite{beery2018recognition,recht2019imagenet} has shown that the performance of neural networks is sensitive to distribution changes and exhibits unsatisfactory performance in new environments.
It is difficult to solve this generalization problem because the observations in the training data cannot cover all real-world environments.
Several kinds of strategies can be applied to tackle OOD generalization~\cite{liu2021towards}.
Causal inference aims to learn causal representations in causal graphs.
By capturing causal representations, the model can obtain potential direct associations, which can help resist distribution changes caused by interventions.
Invariant learning methods represented by invariant risk minimization (IRM)~\cite{arjovsky2019invariant}, which are proposed based on causal inference, have extended generalization models to more practical environments.  
Although the above methods have improved the generalization ability of existing ML models, it is difficult to identify invariance due to the complex topological structure of the graph.
GIL~\cite{li2022learning} captures the invariant relationships within the graph structure and the environment by jointly optimizing three modules.
EERM~\cite{wu2022handling} trains multiple context generators to maximize the variance of risks from multiple virtual environments, to explore invariance within environments.
FLOOD~\cite{liu2023flood} combines invariant representation learning and contrastive learning to train a more flexible framework to deal with different environments.
However, the features captured by the existing invariant learning are relatively generic. Although they can render the model more generalizable, these features may not be beneficial for downstream tasks when dealing with data from other distributions.

\section{Preliminaries}
In this section, we introduce the notation used in this paper, as well as some related methods.
\subsection{Problem Statement} 
The goal of an adversary is to determine whether a given node is used to train a target GNN model or not. 
Formally, 
let $\mathcal{G}=(\mathcal{V}, A, \mathbf{X})$ represents the graph dataset, $\mathcal{V}$ is the node set with size $ n = |\mathcal{V}|$,~and $\mathcal{E} \subseteq \mathcal{V} \times \mathcal{V}$ represents the edge set. We denote the adjacency matrix of $\mathcal{G}$ as $ A \in\{0,1\}^{n \times n}$, where $A_{i j}=1$ if node $v_{i}$ connects to node $v_{j}$, otherwise $A_{i j}=0$, and $\mathcal{N}(v)$ is the neighbor set of node $v$. $\mathbf{X} \in \mathbb{R}^{n \times f}$ is the matrix of node attributes where each row vector $\mathbf{X}_{v} \in \mathbb{R}^{f}$ is the corresponding attributes of node $v$.
Given a target node $v$ in the target dataset $\mathcal{G}_{t}$, a target GNN model $\mathcal{M}_t$, and the adversary’s background knowledge $\mathcal{K}$. 
Membership inference attack $\mathcal{A}$ is defined as:
\begin{equation}
 \mathcal{A}: v, \mathcal{M}_{t}, \mathcal{K} \mapsto\{\text { member, non-membe}\}.
\end{equation}

\subsection{Invariant Learning}
Invariant learning is used to capture invariant relationships among different distributions.
Therefore, when conducting cross-domain attacks, we can utilize invariant learning to capture invariant representations between different datasets.

\noindent \textbf{Empirical risk minimization (ERM)}~\cite{vapnik1991principles} solution is found by minimizing the global risk, expressed as the expected loss over the observational distribution, but it does not generalize well to other domains in the testing\cite{liu2023flood}, ERM as:
\begin{equation}
\mathcal{R}_{\operatorname{ERM}}\left(f_{w}\right)=\mathbb{E}_{P_{\mathrm{obs}}(x, y, e)}\left[\ell\left(f_{w}(x), y\right)\right],
\end{equation}
where $\mathbb{E}_{P_{\mathrm{obs}}(x, y, e)}$ indicates that the expectation is taken with respect to the observed data distribution $P_{\mathrm{obs}}(x, y, e)$, $\ell$ is the loss function and $\omega$ is the parameter.

\noindent \textbf{Invariant Risk Minimization (IRM)}~\cite{arjovsky2019invariant} includes a regularization objective that enables the classifier $f(\cdot )$ to achieve optimality across all environments, following:
\begin{equation}
\mathcal{R}_{\operatorname{IRM}}\left(f_{w}\right)=\sum_{e \in \mathcal{E}^{\text {obs }}} \mathcal{R}_{e}\left(f_{w}\right)+\beta\left\|\nabla_{\omega} \mathcal{R}_{e}\left( f_{\omega}\right)\right\|_{2}^{2}
,\end{equation}
where $\mathcal{E}^{\mathrm{obs}} is$ is the observed environment, $\beta$ is penalty weight, $\mathcal{R}_{e}$ is short for $\mathcal{R}_{\text {ERM}} $ in environment $ e $, and $\left \| \cdot  \right \|_{2}^{2} $ is the square of the $L_{2}$ norm.

\noindent \textbf{Risk Extrapolation (REx)} \cite{krueger2021out} is a form of robust optimization over a perturbation set of extrapolated domains.
That means that reducing differences in risk across training domains can reduce a model’s sensitivity to distribution shifts.
The REx is defined as:
\begin{equation}
\mathcal{R}_{\mathrm{REx}}\left(f_{\omega}\right)=\max _{\substack{\Sigma_{e} \lambda_{e}=1 \\ \lambda_{e} \geq \lambda_{\min }}} \sum_{e \in \mathcal{E}^{\mathrm{obs}}} \lambda_{e} \mathcal{R}_{e}\left(f_{\omega}\right),
\end{equation}
where $\lambda$ is the extrapolated weight. 

\subsection{Graph Information Bottleneck}
The information bottleneck (IB)~\cite{fu2025bi,fu2025discrete} principle uses mutual information $I(X; Y)$ as a cost function and regularization. 
The GIB is defined as follows:
\begin{equation}
\operatorname{GIB}_{\xi }(\mathcal{G}, Y ; Z) \triangleq [-I(Y ; Z)+\xi  I(\mathcal{G} ; Z)],
\end{equation}
where $\xi$ is the Lagrangian parameter to balance the two terms.

\section{Graph Out-Of-Distribution Membership Inference Attack}
In this section, we first define cross-domain MI attacks against GNNs. Then, we discuss the threat model and present the attack methodology.

\begin{figure*}[t]
\centering
\includegraphics[width=1\textwidth]{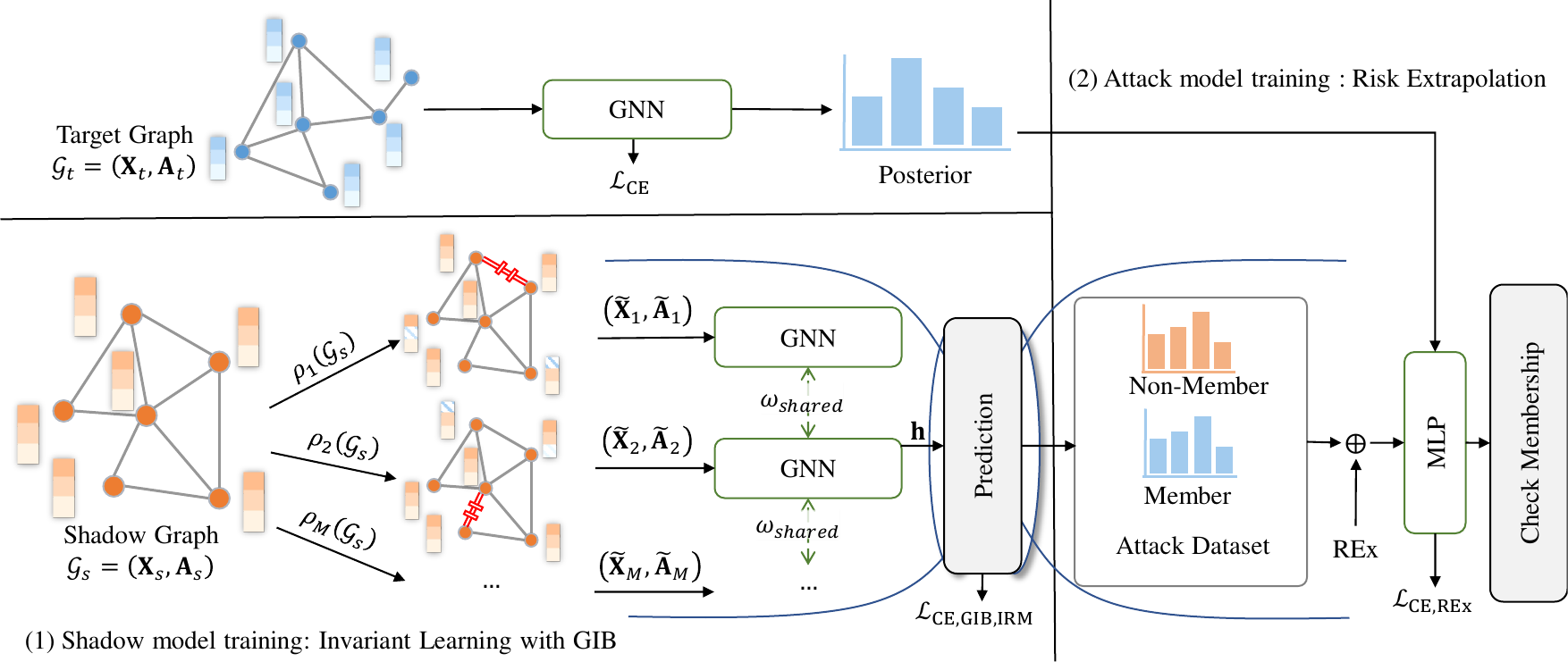} 
\caption
{
Framework of {\modelname}.
(1) The input graph is augmented to construct $M$ training environments. 
Then, a GNN is employed to learn node representations across different environments via Invariant Risk Minimization and Graph Information Bottleneck, aiming to capture the features and structural information in graph data that can be utilized for cross-domain attacks.
Next, (2) the output posteriors of the shadow model are used to construct the attack training set, and variance risk extrapolation is employed to enable the attack model to conduct cross-domain attacks.
}
\label{fig:framework}
\end{figure*}

\subsection{Threat Model}
In this paper, we study MIAs under the black-box settings, which means the adversary can't access the target model’s parameters but can only observe the input and output of the target model, We then analyze the adversary’s background knowledge $\mathcal{K}$ along two dimensions, i.e., shadow dataset, shadow model.

\noindent \textbf{Our setting}.
As mentioned above, it is very difficult to obtain datasets with the same distribution in real life, so our setting is different from previous work~\cite{he2021node,olatunji2021membership}. 
We assume that the attacker uses a dataset with a different distribution from the target dataset for auxiliary training.
Using the shadow dataset, the attacker needs to train a shadow model that can learn invariant features and structures.
However, the generalized representations and structures learned from data in different domains may not fully mimic the target model.
Therefore, based on the above discussion, the purpose of training the shadow model is not only to mimic the behavior of the target model but also to summarize the membership status of data points in the training set of the ML model. 

\subsection{Attack Methodology}
According to the traditional procedure of membership inference attacks on previous ML models and GNN models~\cite{olatunji2021membership,shokri2017membership}, our GOOD-MIA also models the attack model as a binary classification task where the goal is to determine if a given node $v \in V_{t} .$
We illustrate the pipeline of {\modelname} in Fig.~\ref{fig:framework}, which consists of two modules: 
$(1)$ \textbf{Shadow model training}: we adopt IRM and GIB to train the GNN model for invariant representation learning and privacy-sensitive learning. The training environments are constructed by data augmentation on graphs.
$(2)$ \textbf{Attack model training}: we adopt REx to train the binary classification model for OOD generalization. 

\subsubsection{Shadow model training}
To endow the shadow model with generalization ability, we need a shadow model that can capture the invariant properties of graphs, such as features and structures, enabling the posterior distribution learned by the model to be as close as possible to the output posterior of the target model. 
For a shadow dataset $\mathcal{G}_{\text {S }}$, 
the adversary first constructs multiple training environments from the original shadow graph $\mathcal{G}_{\text {S}}$.
Then, each augmented graph $\mathcal{G}_{\text {S }}$ is divided into two disjoint subgraph, including $\mathcal{G}_{\text {S }}^{\text {Train }}$ and $\mathcal{G}_{\text {S }}^{\text {Test }}$.
We perform two typical graph augmentations, namely node feature masking and DropEdge~\cite{you2020graph}:
\begin{equation}
\label{aug}
    \rho_{e}(\mathbf{X}, \mathbf{A})=\left(\widetilde{\mathbf{X}}_{e}, \widetilde{\mathbf{A}}_{e}\right), \quad e=1, \ldots, M,
\end{equation}
where $\widetilde{\mathbf{X}}_{e}$ represents the features after data augmentation, $\widetilde{\mathbf{A}}_{e}$ is the adjacency matrix after data augmentation, and $e$ represents different environments.

Next, we train a GNN encoder $f_{\omega}(\cdot)$ to extract invariant features and to capture information-sensitive representations from the graphs in different training environments.
$f_{\omega}:(\mathrm{X}, \mathrm{~A}) \rightarrow \mathbb{R}^{d}$ is a $L$-layer graph neural networks and outputs $d$-dimension representation for each node.
In layer $l (l = 1,...,L)$, the representation for node $i$ under environment $e$ is defined by:
\begin{equation}
\label{Gnn}
\begin{aligned}
&\mathbf{z}_{e, i}^{(l)}=\operatorname{AGG}\left(\mathbf{h}_{e, i}^{(l-1)},\left\{\mathbf{h}_{e, j}^{(l-1)} \mid j \in \mathcal{N}_{e}(i)\right\}\right),\\
&\mathbf{h}_{e, i}^{(l+1)}=\operatorname{UPDATE}\left(z_{i}^{(l)}, \arg \min _{\mathbf{h}_{e, i}} \mathcal{L}_{\mathrm{GIB}}\right),
\end{aligned}
\end{equation}
where $\mathcal{N}_{e}(i)$ indicates the neighbor set of node $i$ decided by $\widetilde{\mathrm{A}}_{e}$, and $\mathbf{h}_{e, i}^{(0)} = \widetilde{\mathrm{X}}_{e}$.

Finally, a softmax layer is applied to the node representations in the last layer for the final prediction of the node classes. The GNN parameterized by $\omega$ is trained by minimizing the cross-entropy loss defined by:
\begin{equation}
\label{Gnn-loss}
\mathcal{R}_{e}(\omega)=-\frac{1}{N} \sum_{i=1}^{N} \sum_{j=1}^{C} \mathrm{Y}_{i j} \log \left[\sigma\left(f_{\omega}\left(\widetilde{\mathrm{X}}_{e}, \widetilde{\mathrm{~A}}_{e}\right)\right)\right]_{i j},
\end{equation}
where $\sigma$ is the activation function.

To learn a better invariant representation, we use IRM to capture the invariant representations $X_c$ in the graph structure during across different training domains, when $\forall e_{1} \neq e_{2}$,~$P_{e_{1}}(Y \mid X_c)=P_{e_{2}}(Y \mid X_c)$, IRM constructs a linear combination with penalty weights as:
\begin{equation}
\label{IRM}
\begin{aligned}
\mathcal{R}_{\mathrm{IRM}}\left(f_{\omega}\right)=\sum_{e=1}^{M} \mathcal{R}_{e}\left(f_{\omega}\right)+\beta_1\left\|\nabla_{\omega} \mathcal{R}_{e}\left(f_{\omega}\right)\right\|_{2}^{2},
\end{aligned}
\end{equation}
where $\beta_1 \in[0,+\infty)$ controls the balance between reducing average risk and penalty weights of risks.

Overall, in the shadow model training phase, both invariant learning and information bottleneck are jointly optimized under the overall loss as:
\begin{equation}
\label{IRM-IB}
\begin{aligned}
    \min _{\omega}\mathcal{L}_{\text{train}}
    =
    \alpha{\mathrm{GIB}}
    + 
     (1-\alpha)\mathcal{R}_{\mathrm{IRM}},
\end{aligned}
\end{equation}
where $\alpha \in[0,1)$ is the weight factor used to balance the constant risk and the graph information bottleneck.
\subsubsection{Attack model training}
The attack model is a binary machine learning classifier and its input is derived from a node’s posteriors provided by a GNN.
The MLP parameterized by $(w, b)$ is trained by minimizing the cross-entropy loss defined as.
\begin{equation}
\label{att-loss}
    \mathcal{R}_{e}(w, b)=-\frac{1}{N} \sum_{i=1}^{N} \sum_{j=1}^{C} \mathbf{y}_{i j} \log (p_{i j}).
\end{equation}

To enable the attack model to obtain a good generalization ability of the attack, we adopt the REx principle.
The goal of using REx is to reduce the differences in risks across different domains, thereby enhancing the model's robustness against distribution shifts. 
It encourages the equality of training risks and when a distribution shift occurs at test time, the risks are more likely to change less.
Minimax-REx builds a linear affine combination of training risks, as represented by:
\begin{equation}
\label{MM-REx}
\begin{aligned}
    \mathcal{R}_{\mathrm{MM}-\mathrm{REx}}(\psi, b) \doteq &\max _{\substack{\Sigma_{e} \lambda_{e}=1 \\ \lambda_{e} \geq \lambda_{\min }}} \sum_{e=1}^{M} \lambda_{e} \mathcal{R}_{e}(\psi, b)
    \\&=\left(1-M \lambda_{\min }\right) \max _{e} \mathcal{R}_{e}(\psi, b)
    \\&+\lambda_{\min } \sum_{e=1}^{M} \mathcal{R}_{e}(\psi, b),
\end{aligned}
\end{equation}
where $M$ is the number of environments, which consists of a posteriori of the enhanced graph.

Practically, as there is a maximum in Eq.~\eqref{MM-REx}, it is hard and unstable to optimize $\mathcal{R}_{\mathrm{MM}-\mathrm{REx}}$. 
To tackle the problem, we replace Eq.~\eqref{MM-REx} with the variance of risks 
as: 

\begin{equation}
\label{V-REx}
\begin{aligned}
   \mathcal{R}_{\mathrm{V}-\mathrm{REx}}(\psi, b) 
   &\doteq \beta_2 \operatorname{Var}\left(\left\{\mathcal{R}_{1}(\psi, b), \ldots, \mathcal{R}_{M}(\psi, b)\right\}\right)
   \\&+\sum_{e=1}^{M} \mathcal{R}_{e}(\psi, b),
\end{aligned}
\end{equation}
where $\beta_2 \in[0,+\infty)$ controls the balance between reducing average risk and enforcing equality of risks.

\subsection{Overall Algorithm and Complexity Analysis}
The overall training algorithm is summarized in Algorithm \ref{alg:algorithm}.
Given a shadow graph $\mathcal{G}_{\text {S}}$, we first construct $M$ training environments by data augmentation (Line 1).
With the GNN initialized parameters (Line 2), we get the node representation under each environment (Line 5).
After that, the model is trained until convergence by minimizing Eq.(\ref{Gnn-loss}) (Line 7).
In the attack model training phase, We use the output posterior of the shadow model to construct the attack dataset and regard the posterior outputs of different augmented graphs as different environments in the attack dataset.
Finally, the attack model is updated by minimizing Eq. (\ref{V-REx}) (Line 11).

Consider a graph with $N$ nodes and $E$ edges, the average degree is $\bar{d}$, an MLP with $N$ samples, $D$ input dimensions, and $H$ dimensions of the hidden layer. GNN with $L$ layers computes embeddings in time $O\left(NL \bar{d}^{2}\right)$ and the time complexity of the MLP is $O\left(NDH\right)$.
{\modelname}~does $M$ encoder computations per update step ($M$ for training environment) plus a prediction step.
The overall time complexity has a linear relationship with the previous work.

\begin{algorithm}[tb]
    \caption{GOOD-MIA: Out-of-Distribution Membership Inference Attack Approach for Cross-domain Graphs Attack}
    \label{alg:algorithm}
    \textbf{Input}: target graph $\mathcal{G}_{\text {target}}$, shadow graph $\mathcal{G}_{\text {shadow}}$, Number of training environments $M$\\
    \textbf{Parameter}: $\omega $, $\psi $\\
    \textbf{Output}: Membership prediction
    
    \begin{algorithmic}[1] 
        \STATE Construct $M$ shadow training environments by Eq.~(\ref{aug}); 
        \STATE Initialization parameters $\omega $, $\psi $;
        \WHILE{Shadow model training epoch $<$ $N_{epoch}^{S}$}
        \FOR{$e = 1,\ldots,M$}
        \STATE Get the representation $\textit{\bm{\mathbf{h}}}$ for nodes of $\mathcal{G}_{e}$ w.r.t Eq.~(\ref{Gnn});
        \ENDFOR
        \STATE Train $\omega $ by minimizing Eq.~(\ref{IRM-IB});
        \ENDWHILE
        \STATE \textbf{return} Posterior probability for shadow model;
        \WHILE{Attack model training epoch $<$ $N_{epoch}^{Attack}$}
        \STATE Train $\omega $ by minimizing Eq.~(\ref{V-REx});
        \ENDWHILE
    \end{algorithmic}
\end{algorithm}

\begin{table}[h]
\scalebox{0.85}{
    \centering
    \begin{tabular}{c|c|c|cc}
        \toprule
        Datasets & \#Nodes & \#Edges & \#Classes & \#Features \\
        \midrule
        Cora & 2708 & 5429 & 7 & 1433 \\
        Citeseer & 3327 & 4732 & 6 & 3703 \\
        Pubmed & 19717 & 44338 & 3 & 500 \\
        \midrule 
        Twitch & 1912 - 9498 & 31299 - 153138 & 2 & 2545 \\
        FB-100 & 769 - 41536 & 16656 - 1590655 & 2 & 8319 \\
        \bottomrule
    \end{tabular}
    }
    \caption{Statistics for experimental datasets}
    \label{tab:dataset}
\end{table}

\begin{table*}[ht]
\centering
\begin{tabular}{cc|ccc|ccc|ccc}
\toprule
&\multirow{2}{*}{\raisebox{-1ex}{Model}}
& \multicolumn{3}{c|}{Cora} 
& \multicolumn{3}{c|}{CiteSeer} 
& \multicolumn{3}{c}{Pubmed}
\\ \cmidrule{3-11}
&
&  ACC &  AUC & Recall 
&  ACC &  AUC & Recall 
&  ACC &  AUC & Recall

\\ \midrule
\multirow{4}{*}{{TSTS}} &
GCN    &67.31   &67.34   &67.18   &79.34   &79.39   &78.35 &50.57 &50.58 &50.60\\
&GAT   &65.81   &65.33   &65.83   &77.49   &77.57   &77.96 &50.18 &50.19 &50.18 \\
&SGC   &66.36   &66.28   &66.33   &80.21   &81.52   &80.09 &51.19 &51.22 &51.19  \\

\midrule
\multirow{4}{*}{\textbf{\modelname}} &
GCN   &\textbf{74.15}  &\textbf{73.99}  &\textbf{73.02}   &\textbf{84.10}  &\textbf{84.04}   &\underline{84.26} &53.79   &53.81   &53.79  \\
&GAT  &\underline{71.23}  &71.14  &70.99   &81.81  &82.48   &81.87 &\textbf{54.47}   &\textbf{54.39}   &\textbf{54.48}  \\
&SGC  &71.07  &\underline{71.80}   &\underline{71.40}   &\underline{82.72}  &\underline{82.83}   &\textbf{84.67} &\underline{53.94}   &\underline{53.92}   &\underline{53.95}  \\

\midrule
\multirow{3}{*}{{Ablation}} &
{\modelname}\textbackslash IRM  &62.87   &62.96   &62.87  &62.33   &62.34   &62.33   &45.72   &45.76   &45.73 \\
&{\modelname}\textbackslash GIB &68.23   &68.35   &68.25  &68.53   &69.28   &70.50   &48.20   &47.20   &47.26 \\
&{\modelname}\textbackslash REx &61.49   &61.45   &61.50  &71.02   &70.90   &71.30   &51.73   &51.71   &51.73  \\
\bottomrule
\end{tabular}
\caption{Summary results of Synthetic Data~(\textbf{Bold}: best;~\underline{Underline}: runner-up).}
\label{tab:Synthetic Data}
\end{table*}

\section{Experiments}
In this section, we investigate the effectiveness of the proposed attack model in the face of three cross-domain settings with practical significance, aiming to address the following research questions.

\subsection{Experimental settings}
\textbf{Datasets.}
We adopt five node property prediction datasets of different sizes and properties, including Cora, Citeseer, Pubmed, Twitch and Facebook-100.
For Cora, Citeseer and Pubmed~\cite{sen2008collective}, we keep the original node labels and synthetically create spurious node features to introduce distribution shifts between different domain data.
The Twitch and Facebook-100~\cite{rozemberczki2021twitch,traud2012social} represent canonical real-world social networks.
For Twitch, we consider subgraph-level data splits: 
nodes in subgraph DE are used as target model datasets, while nodes in ENGB, ES, FR, PTBR, RU and TW are used as shadow model datasets to set cross-domain attacks in different domain environments.
For Facebook-100, we use John Hopkins, Amherst and Cornell5 as the target datasets, Penn and Reed as the shadow datasets.
We summarize the dataset information in Tab.~\ref{tab:dataset}.

\noindent \textbf{Baselines.} We established three backbone models (GCN, GAT, SGC) to test the expressiveness of {\modelname} in different scenarios.
We compare the proposed~{\modelname} with the state-of-the-art attack methods TSTS~\cite{olatunji2021membership} on GNNs.
The baselines in our experiment have the same setting as our method, that is, the shadow dataset has a different distribution from the target dataset.

\noindent \textbf{Evaluation Metrics.}
We use Accuracy, AUC score, and Recall~\cite{he2021node,olatunji2021membership,salem2018ml} to evaluate the performance of the attack model.

\begin{table*}[ht]
\scalebox{0.97}{
\centering
\begin{tabular}{cc|cc|cc|cc|cc|cc|cc}
\toprule
&\multirow{2}{*}{\raisebox{-1ex}{Model}}
& \multicolumn{2}{c}{ENGB} 
& \multicolumn{2}{c}{ES} 
& \multicolumn{2}{c}{FR}
& \multicolumn{2}{c}{PTBR}
& \multicolumn{2}{c}{RU}
& \multicolumn{2}{c}{TW}

\\ \cmidrule{3-14}
&
&  ACC &  AUC &  ACC &  AUC &  ACC &  AUC
&  ACC &  AUC &  ACC &  AUC &  ACC &  AUC

\\ \midrule
\multirow{3}{*}{{TSTS}} &
GCN    &57.04   &56.99   &59.65   &59.67   &59.07   &59.08 &58.99   &59.07   &56.59   &56.56   &50.05   &60.08  \\
&GAT   &55.69   &55.67   &56.81   &56.78   &56.31   &55.32 &57.37   &57.36   &57.33   &57.39   &56.14   &56.18  \\
&SGC   &\underline{60.27}   &\underline{60.20}   &59.43   &59.47   &60.14   &60.20 &\underline{61.37}   &\underline{61.36}   &58.21   &58.27   &60.15   &60.14  \\

\midrule
\multirow{3}{*}{\textbf{\modelname}} &
GCN    &57.14   &57.06   &\underline{61.06}   &\underline{61.10}   &\underline{63.02}   &\underline{63.15} &61.02   &61.04   &58.88   &58.92   &\underline{61.30}   &\underline{61.27}  \\
&GAT   &59.97   &60.00   &59.46   &59.50   &59.29   &59.30 &59.52   &59.54   &\underline{58.94}   &\underline{58.96}   &60.12   &60.19  \\
&SGC   &\textbf{61.29}  &\textbf{61.31}   &\textbf{61.56}   &\textbf{61.58}   &\textbf{63.74}   &\textbf{63.77} &\textbf{64.80}   &\textbf{64.92}   &\textbf{60.38}   &\textbf{60.36}   &\textbf{61.56}   &\textbf{61.33} \\

\bottomrule
\end{tabular}
}
\caption{Attack accuracy and AUC of the~{\modelname}~on the Twitch dataset~(\textbf{Bold}: best;~\underline{Underline}: runner-up).}
\label{tab:Twitch}
\end{table*}

\begin{table*}[ht]
\centering
\scalebox{0.87}{
\centering
\begin{tabular}{c|ccc|ccc|ccc|ccc}
\toprule
\multirow{3}{*}{{\diagbox{Target Data}{Shadow Data}}}
&\multicolumn{6}{c|}{Penn} 
& \multicolumn{6}{c}{Reed} 
\\ \cmidrule{2-13}
&\multicolumn{3}{c}{\modelname} &\multicolumn{3}{c|}{TSTS} &\multicolumn{3}{c}{\modelname} &\multicolumn{3}{c}{TSTS}\\
\cmidrule{2-13}
&  ACC &  AUC & Recall &  ACC &  AUC & Recall &  ACC &  AUC & Recall &  ACC &  AUC & Recall
\\ \midrule
John Hopkins & \textbf{57.20}    & 57.12   & \textbf{57.20}     & 53.39  & 53.42  & 53.39  & 57.74   & \textbf{57.75}   & 57.73    & 51.99  & 51.97  & 51.99  \\
\midrule
Amherst & \textbf{58.31}   & 57.83   & \textbf{58.31}    & 51.62  & 51.63  & 51.62  & 60.19   & \textbf{60.29}   & 57.68    & 54.56  & 54.48  & 54.56  \\
\midrule
Cornell5     & 57.52   & \textbf{57.59}   & 57.53    & 53.68  & 53.62  & 53.62  & 56.81   & 56.78   & \textbf{56.83}    & 53.48  & 53.47  & 53.48   \\ 
\bottomrule
\end{tabular}
}
\caption{Attack accuracy and AUC of the~{\modelname} on the Facebook-100 dataset~(\textbf{Bold}: best).}
\label{tab:Facebook}
\end{table*}
\subsection{Performance Evaluation}
We conduct comprehensive performance verification and ablation experiments on~{\modelname}.

\noindent \textbf{Distribution Shifts on Synthetic Data.}
We report the attack scores on the citation network dataset in Tab.~\ref{tab:Synthetic Data}.
We found that when using different GNNs as the backbone, {\modelname} consistently showed a significant advantage over corresponding competitors under artificially induced distribution shifts.
Also, when the datasets were Cora and PubMed, their performance was close to that of the attack models trained with datasets of the same distribution~\cite{olatunji2021membership}.
The results indicate that this attack model can effectively capture cross-domain features and their structures when using datasets with different distributions, enabling the model to launch effective attacks.

\noindent \textbf{Distribution Shifts across domains.}
Tab.~\ref{tab:Twitch} and Tab.~\ref{tab:Facebook} demonstrate the attack performance in real-world social network datasets.
This kind of dataset has great practical significance because it is rather difficult for us to obtain datasets with the same distribution, while it is relatively easy to acquire datasets with approximate distributions.
This kind of dataset is challenging for cross-domain attacks since the nodes in different subgraphs are disconnected.
We found {\modelname} achieves overall superior performance over competitors.
This demonstrates the efficacy of our model in tackling OOD attacks across graphs in different domains.

\noindent \textbf{Overall.} 
For different datasets, the invariance of their features and structures may vary in strength.  
For example, citation networks may exhibit strong invariance in their structures, making them easier to capture in cross-domain scenarios.  
Social networks may exhibit weak invariance in their structures, leading to less clear capture and, consequently, poorer attack performance.

\subsection{Ablation Study}
In this section, we analyze the effectiveness of the three variants:
\begin{itemize}
\item \textbf{{\modelname} ($\bm{w / o\,IRM}$ ):}~We remove the IRM in the shadow model training objective (Eq.~\eqref{IRM-IB}).
\item \textbf{{\modelname} ($\bm{w / o\,GIB}$ ):} We remove the GIB in the shadow model training objective (Eq.~\eqref{IRM-IB}).
\item \textbf{{\modelname} ($\bm{w / o\,REx}$ ):} We remove the REx in the attack model training objective (Eq.~\eqref{V-REx}), and using the traditional Cross Entropy Loss, and only with the IRM and GIB in the overall training.
\end{itemize}
We choose GCN as the backbone to evaluate three key modules of GOOD-MIA, namely IRM, GIB, and REx, by removing each module respectively.
We report the ablation experiments of artificial data MIA in Tab.~\ref{tab:Synthetic Data}.
Compared with the three variants, the complete model achieves the best attack performance in terms of the total number, indicating that each module is essential for the generalization of the attack model.

Removing the IRM part, when training the shadow model, {\modelname}\textbackslash IRM can only capture features and structures related to downstream tasks through the information bottleneck.  
It fails to obtain invariant representations of the data when the distributions vary.  
Merely conducting risk extrapolation via the attack model cannot yield a satisfactory attack effect.

Omitting the GIB part, {\modelname}\textbackslash GIB can only obtain generalized representations of data during shadow model training.   Although such information may have the same probability under different distributions, it may not be a crucial part of downstream tasks in different distributions.

Without the REx part, although {\modelname}\textbackslash REx can capture invariant representations as well as structures and features related to downstream tasks during the training of the shadow model, simply using a simple MLP is not sufficient to extrapolate the posterior features learned by the shadow model to other data domains.
We conclude that only by combining these three can we improve the capability of cross-domain attacks.

\subsection{Analysis}
\begin{figure}[t]
\centering
\includegraphics[width=0.9\columnwidth]{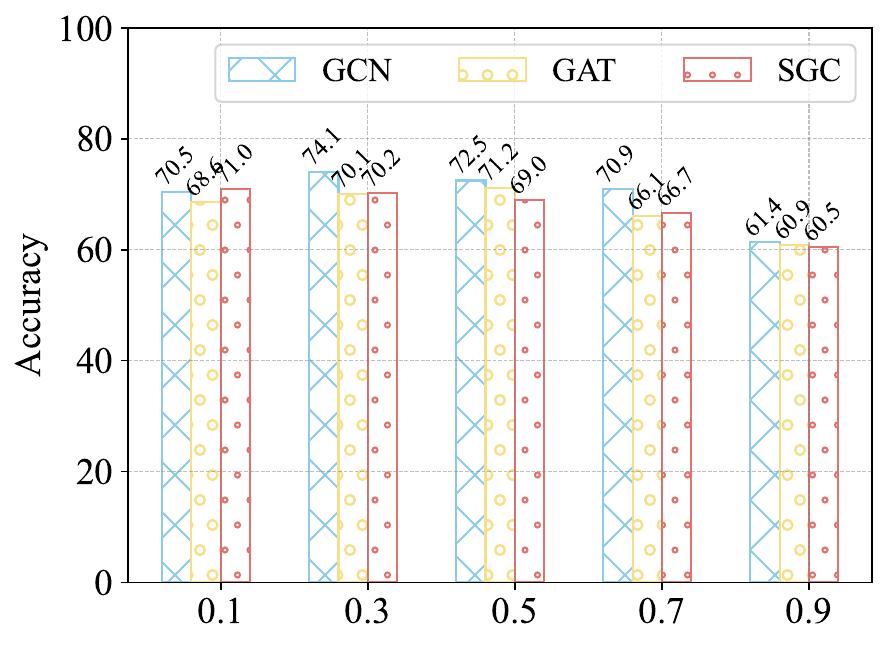}
\centering
\caption{Trade-off parameter $\alpha$ analysis.}
\label{fig:hyperpara}
\end{figure}

\begin{figure}[t]
\centering
\includegraphics[width=0.9\columnwidth]{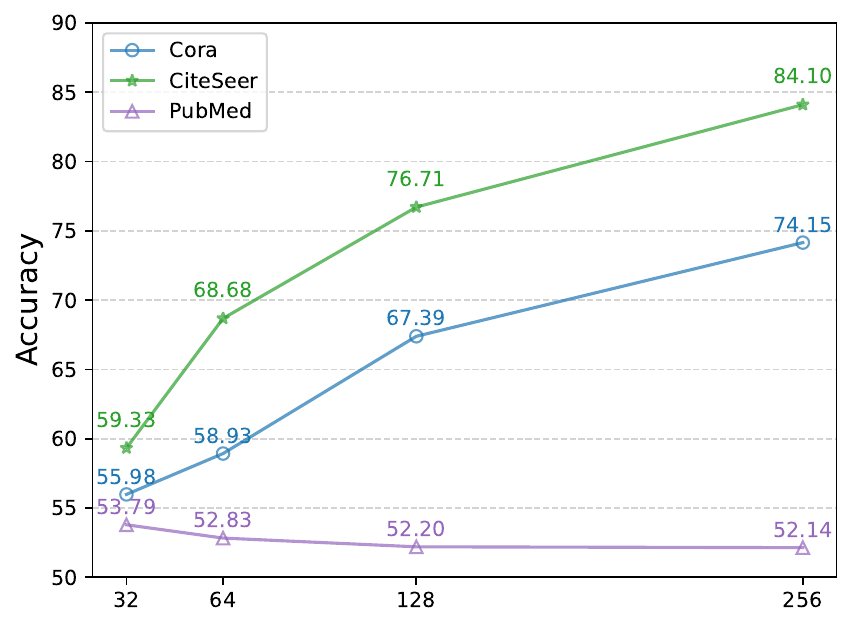} 
\centering
\caption{Different numbers of neurons are used in the hidden layer of the shadow model for the MIA.} 

\label{fig:NumberofNeurons}
\end{figure}
\textbf{Hyperparameter Trade-off Analysis.}~We conduct a hyperparameter analysis of the trade-off parameter in the shadow model to verify the roles of information bottleneck and invariant learning in {\modelname}. 
The results are shown in Fig.~\ref{fig:hyperpara}, indicating that the sensitivity to attack accuracy when setting $\alpha$ varies between different GNN models.
It can be seen from this that the analysis we presented earlier is correct.
When the shadow model learns invariant representations using IRM, generalizing to other domains may not be relevant to the downstream tasks.
Therefore, it is also necessary to capture the features and structures that are closely related to the downstream tasks.
Moreover, when using different GNN models, the attack effect obtained by using different parameters is also different due to different aggregation methods.

In summary, the hyperparameter settings can be further optimized based on the characteristics of different models to improve the overall performance of the attack model.

\noindent \textbf{Different Model Architecture.}
We further investigate whether the different number of neurons affects the attack performance.
We evaluated the target model with 256 neurons in its hidden layer, while the number of neurons in the shadow model varied from 32 to 256.
The results are depicted in Fig.~\ref {fig:NumberofNeurons}.
We observed that in Cora and CiteSeer, the closer the shadow model approximates the target model, the higher the attack accuracy.
PubMed is the opposite.
This may be attributed to the lower feature dimensionality of PubMed.

\section{Conclusions}
In this paper, we propose a novel framework named~{\modelname}, designed to explore the feasibility of cross-domain membership inference attacks when identically distributed auxiliary datasets are unavailable in real-world scenarios, and to enhance the effectiveness of such cross-domain attacks.

We decompose the overall objectives of~{\modelname} into shadow model training and attack model training, which have a linear relationship. 
During the shadow model training phase, invariant features and key graph structures are first captured from the environments of different graph data. Subsequently, attack training sets for different environments are constructed, followed by training the attack model.
Finally, through risk extrapolation, the attack model can be generalized to other domains for attacks. 
Extensive experiments show that~{\modelname} exhibits excellent attack inference capabilities and domain adaptability.

\section*{Acknowledgments}
The corresponding authors are Xingcheng Fu and Yuecen Wei. This paper was supported by the National Natural Science Foundation of China (Nos. 62162005, 62462007 and U21A20474), National Natural Science Foundation Joint Cultivation Project of Guangxi Normal University (No. 2024PY028), Guangxi Bagui Youth Talent Training Program, Guangxi Collaborative Innovation Center of Multisource Information Integration and Intelligent Processing and the Key Lab of Education Blockchain and Intelligent Technology, Ministry of Education (EBME24-01).

\section*{Contribution Statement}
*Co-corresponding Authors.

\bibliographystyle{named}
\bibliography{references}

\begin{thebibliography}{}

\bibitem[\protect\citeauthoryear{Arjovsky \bgroup \em et al.\egroup }{2019}]{arjovsky2019invariant}
Martin Arjovsky, L{\'e}on Bottou, Ishaan Gulrajani, and David Lopez-Paz.
\newblock Invariant risk minimization.
\newblock {\em arXiv preprint arXiv:1907.02893}, 2019.

\bibitem[\protect\citeauthoryear{Beery \bgroup \em et al.\egroup }{2018}]{beery2018recognition}
Sara Beery, Grant Van~Horn, and Pietro Perona.
\newblock Recognition in terra incognita.
\newblock In {\em Proceedings of the European conference on computer vision (ECCV)}, pages 456--473, 2018.

\bibitem[\protect\citeauthoryear{Boll \bgroup \em et al.\egroup }{2024}]{boll2024graph}
Helo{\'\i}sa~Oss Boll, Ali Amirahmadi, Mirfarid~Musavian Ghazani, Wagner~Ourique de~Morais, Edison~Pignaton de~Freitas, Amira Soliman, Farzaneh Etminani, Stefan Byttner, and Mariana Recamonde-Mendoza.
\newblock Graph neural networks for clinical risk prediction based on electronic health records: A survey.
\newblock {\em J. Biomed. Informatics}, 151:104616, 2024.

\bibitem[\protect\citeauthoryear{Fu \bgroup \em et al.\egroup }{2023}]{fu2023hyperbolic}
Xingcheng Fu, Yuecen Wei, Qingyun Sun, Haonan Yuan, Jia Wu, Hao Peng, and Jianxin Li.
\newblock Hyperbolic geometric graph representation learning for hierarchy-imbalance node classification.
\newblock In {\em Proceedings of the ACM Web Conference 2023}, pages 460--468, 2023.

\bibitem[\protect\citeauthoryear{Fu \bgroup \em et al.\egroup }{2024}]{fu2024hyperbolic}
Xingcheng Fu, Yisen Gao, Yuecen Wei, Qingyun Sun, Hao Peng, Jianxin Li, and Xianxian Li.
\newblock Hyperbolic geometric latent diffusion model for graph generation.
\newblock {\em arXiv preprint arXiv:2405.03188}, 2024.

\bibitem[\protect\citeauthoryear{Fu \bgroup \em et al.\egroup }{2025a}]{fu2025bi}
Xingcheng Fu, Yisen Gao, Beining Yang, Yuxuan Wu, Haodong Qian, Qingyun Sun, and Xianxian Li.
\newblock Bi-directional multi-scale graph dataset condensation via information bottleneck.
\newblock In {\em Proceedings of the AAAI Conference on Artificial Intelligence}, volume~39, pages 16674--16681, 2025.

\bibitem[\protect\citeauthoryear{Fu \bgroup \em et al.\egroup }{2025b}]{fu2025discrete}
Xingcheng Fu, Jian Wang, Yisen Gao, Qingyun Sun, Haonan Yuan, Jianxin Li, and Xianxian Li.
\newblock Discrete curvature graph information bottleneck.
\newblock In {\em Proceedings of the AAAI Conference on Artificial Intelligence}, volume~39, pages 16666--16673, 2025.

\bibitem[\protect\citeauthoryear{Hayes \bgroup \em et al.\egroup }{2017}]{hayes2017logan}
Jamie Hayes, Luca Melis, George Danezis, and Emiliano De~Cristofaro.
\newblock Logan: Membership inference attacks against generative models.
\newblock {\em arXiv preprint arXiv:1705.07663}, 2017.

\bibitem[\protect\citeauthoryear{He \bgroup \em et al.\egroup }{2020}]{he2020segmentations}
Yang He, Shadi Rahimian, Bernt Schiele, and Mario Fritz.
\newblock Segmentations-leak: Membership inference attacks and defenses in semantic image segmentation.
\newblock In {\em Computer Vision--ECCV 2020: 16th European Conference, Glasgow, UK, August 23--28, 2020, Proceedings, Part XXIII 16}, pages 519--535. Springer, 2020.

\bibitem[\protect\citeauthoryear{He \bgroup \em et al.\egroup }{2021}]{he2021node}
Xinlei He, Rui Wen, Yixin Wu, Michael Backes, Yun Shen, and Yang Zhang.
\newblock Node-level membership inference attacks against graph neural networks.
\newblock {\em arXiv preprint arXiv:2102.05429}, 2021.

\bibitem[\protect\citeauthoryear{Hintersdorf \bgroup \em et al.\egroup }{2021}]{hintersdorf2021trust}
Dominik Hintersdorf, Lukas Struppek, and Kristian Kersting.
\newblock To trust or not to trust prediction scores for membership inference attacks.
\newblock {\em arXiv preprint arXiv:2111.09076}, 2021.

\bibitem[\protect\citeauthoryear{Krueger \bgroup \em et al.\egroup }{2021}]{krueger2021out}
David Krueger, Ethan Caballero, Joern-Henrik Jacobsen, Amy Zhang, Jonathan Binas, Dinghuai Zhang, Remi Le~Priol, and Aaron Courville.
\newblock Out-of-distribution generalization via risk extrapolation (rex).
\newblock In {\em International conference on machine learning}, pages 5815--5826. PMLR, 2021.

\bibitem[\protect\citeauthoryear{Li \bgroup \em et al.\egroup }{2022}]{li2022learning}
Haoyang Li, Ziwei Zhang, Xin Wang, and Wenwu Zhu.
\newblock Learning invariant graph representations for out-of-distribution generalization.
\newblock {\em Advances in Neural Information Processing Systems}, 35:11828--11841, 2022.

\bibitem[\protect\citeauthoryear{Li \bgroup \em et al.\egroup }{2025}]{li2025rethinking}
Xianxian Li, Zeming Gan, Qiyu Li, Bin Qu, Jinyan Wang, et~al.
\newblock Rethinking the impact of noisy labels in graph classification: A utility and privacy perspective.
\newblock {\em Neural Networks}, 182:106919, 2025.

\bibitem[\protect\citeauthoryear{Liu \bgroup \em et al.\egroup }{2021}]{liu2021towards}
Jiashuo Liu, Zheyan Shen, Yue He, Xingxuan Zhang, Renzhe Xu, Han Yu, and Peng Cui.
\newblock Towards out-of-distribution generalization: A survey.
\newblock {\em arXiv preprint arXiv:2108.13624}, 2021.

\bibitem[\protect\citeauthoryear{Liu \bgroup \em et al.\egroup }{2023}]{liu2023flood}
Yang Liu, Xiang Ao, Fuli Feng, Yunshan Ma, Kuan Li, Tat-Seng Chua, and Qing He.
\newblock Flood: A flexible invariant learning framework for out-of-distribution generalization on graphs.
\newblock In {\em Proceedings of the 29th ACM SIGKDD conference on knowledge discovery and data mining}, pages 1548--1558, 2023.

\bibitem[\protect\citeauthoryear{Olatunji \bgroup \em et al.\egroup }{2021}]{olatunji2021membership}
Iyiola~E Olatunji, Wolfgang Nejdl, and Megha Khosla.
\newblock Membership inference attack on graph neural networks.
\newblock In {\em 2021 Third IEEE International Conference on Trust, Privacy and Security in Intelligent Systems and Applications (TPS-ISA)}, pages 11--20. IEEE, 2021.

\bibitem[\protect\citeauthoryear{Recht \bgroup \em et al.\egroup }{2019}]{recht2019imagenet}
Benjamin Recht, Rebecca Roelofs, Ludwig Schmidt, and Vaishaal Shankar.
\newblock Do imagenet classifiers generalize to imagenet?
\newblock In {\em International conference on machine learning}, pages 5389--5400. PMLR, 2019.

\bibitem[\protect\citeauthoryear{Rozemberczki and Sarkar}{2021}]{rozemberczki2021twitch}
Benedek Rozemberczki and Rik Sarkar.
\newblock Twitch gamers: a dataset for evaluating proximity preserving and structural role-based node embeddings.
\newblock {\em arXiv preprint arXiv:2101.03091}, 2021.

\bibitem[\protect\citeauthoryear{Salem \bgroup \em et al.\egroup }{2018}]{salem2018ml}
Ahmed Salem, Yang Zhang, Mathias Humbert, Pascal Berrang, Mario Fritz, and Michael Backes.
\newblock Ml-leaks: Model and data independent membership inference attacks and defenses on machine learning models.
\newblock {\em arXiv preprint arXiv:1806.01246}, 2018.

\bibitem[\protect\citeauthoryear{Sen \bgroup \em et al.\egroup }{2008}]{sen2008collective}
Prithviraj Sen, Galileo Namata, Mustafa Bilgic, Lise Getoor, Brian Galligher, and Tina Eliassi-Rad.
\newblock Collective classification in network data.
\newblock {\em AI magazine}, 29(3):93--93, 2008.

\bibitem[\protect\citeauthoryear{Sharma \bgroup \em et al.\egroup }{2024}]{sharma2024survey}
Kartik Sharma, Yeon-Chang Lee, Sivagami Nambi, Aditya Salian, Shlok Shah, Sang-Wook Kim, and Srijan Kumar.
\newblock A survey of graph neural networks for social recommender systems.
\newblock {\em ACM Computing Surveys}, 56(10):1--34, 2024.

\bibitem[\protect\citeauthoryear{Shokri \bgroup \em et al.\egroup }{2017}]{shokri2017membership}
Reza Shokri, Marco Stronati, Congzheng Song, and Vitaly Shmatikov.
\newblock Membership inference attacks against machine learning models.
\newblock In {\em 2017 IEEE symposium on security and privacy (SP)}, pages 3--18. IEEE, 2017.

\bibitem[\protect\citeauthoryear{Song and Shmatikov}{2019}]{song2019auditing}
Congzheng Song and Vitaly Shmatikov.
\newblock Auditing data provenance in text-generation models.
\newblock In {\em Proceedings of the 25th ACM SIGKDD International Conference on Knowledge Discovery \& Data Mining}, pages 196--206, 2019.

\bibitem[\protect\citeauthoryear{Traud \bgroup \em et al.\egroup }{2012}]{traud2012social}
Amanda~L Traud, Peter~J Mucha, and Mason~A Porter.
\newblock Social structure of facebook networks.
\newblock {\em Physica A: Statistical Mechanics and its Applications}, 391(16):4165--4180, 2012.

\bibitem[\protect\citeauthoryear{Tu \bgroup \em et al.\egroup }{2021}]{tu2021deep}
Wenxuan Tu, Sihang Zhou, Xinwang Liu, Xifeng Guo, Zhiping Cai, En~Zhu, and Jieren Cheng.
\newblock Deep fusion clustering network.
\newblock In {\em Proceedings of the AAAI conference on artificial intelligence}, volume~35, pages 9978--9987, 2021.

\bibitem[\protect\citeauthoryear{Vapnik}{1991}]{vapnik1991principles}
Vladimir Vapnik.
\newblock Principles of risk minimization for learning theory.
\newblock {\em Advances in neural information processing systems}, 4, 1991.

\bibitem[\protect\citeauthoryear{Veli{\v{c}}kovi{\'c} \bgroup \em et al.\egroup }{2017}]{velivckovic2017graph}
Petar Veli{\v{c}}kovi{\'c}, Guillem Cucurull, Arantxa Casanova, Adriana Romero, Pietro Lio, and Yoshua Bengio.
\newblock Graph attention networks.
\newblock {\em arXiv preprint arXiv:1710.10903}, 2017.

\bibitem[\protect\citeauthoryear{Wei \bgroup \em et al.\egroup }{2024}]{wei2024poincare}
Yuecen Wei, Haonan Yuan, Xingcheng Fu, Qingyun Sun, Hao Peng, Xianxian Li, and Chunming Hu.
\newblock Poincar{\'e} differential privacy for hierarchy-aware graph embedding.
\newblock In {\em Proceedings of the AAAI Conference on Artificial Intelligence}, volume~38, pages 9160--9168, 2024.

\bibitem[\protect\citeauthoryear{Wei \bgroup \em et al.\egroup }{2025}]{wei2025prompt}
Yuecen Wei, Xingcheng Fu, Lingyun Liu, Qingyun Sun, Hao Peng, and Chunming Hu.
\newblock Prompt-based unifying inference attack on graph neural networks.
\newblock In {\em Proceedings of the AAAI Conference on Artificial Intelligence}, volume~39, pages 12836--12844, 2025.

\bibitem[\protect\citeauthoryear{Wu \bgroup \em et al.\egroup }{2019}]{wu2019simplifying}
Felix Wu, Amauri Souza, Tianyi Zhang, Christopher Fifty, Tao Yu, and Kilian Weinberger.
\newblock Simplifying graph convolutional networks.
\newblock In {\em International conference on machine learning}, pages 6861--6871. Pmlr, 2019.

\bibitem[\protect\citeauthoryear{Wu \bgroup \em et al.\egroup }{2020a}]{wu2020graph}
Tailin Wu, Hongyu Ren, Pan Li, and Jure Leskovec.
\newblock Graph information bottleneck.
\newblock {\em Advances in Neural Information Processing Systems}, 33:20437--20448, 2020.

\bibitem[\protect\citeauthoryear{Wu \bgroup \em et al.\egroup }{2020b}]{wu2020comprehensive}
Zonghan Wu, Shirui Pan, Fengwen Chen, Guodong Long, Chengqi Zhang, and Philip~S Yu.
\newblock A comprehensive survey on graph neural networks.
\newblock {\em IEEE transactions on neural networks and learning systems}, 32(1):4--24, 2020.

\bibitem[\protect\citeauthoryear{Wu \bgroup \em et al.\egroup }{2022}]{wu2022handling}
Qitian Wu, Hengrui Zhang, Junchi Yan, and David Wipf.
\newblock Handling distribution shifts on graphs: An invariance perspective.
\newblock {\em arXiv preprint arXiv:2202.02466}, 2022.

\bibitem[\protect\citeauthoryear{You \bgroup \em et al.\egroup }{2020}]{you2020graph}
Yuning You, Tianlong Chen, Yongduo Sui, Ting Chen, Zhangyang Wang, and Yang Shen.
\newblock Graph contrastive learning with augmentations.
\newblock {\em Advances in neural information processing systems}, 33:5812--5823, 2020.

\bibitem[\protect\citeauthoryear{Zhang \bgroup \em et al.\egroup }{2021}]{zhang2021graph}
Xiao-Meng Zhang, Li~Liang, Lin Liu, and Ming-Jing Tang.
\newblock Graph neural networks and their current applications in bioinformatics.
\newblock {\em Frontiers in genetics}, 12:690049, 2021.

\bibitem[\protect\citeauthoryear{Zhang \bgroup \em et al.\egroup }{2024a}]{zhang2024bayesian}
Guixian Zhang, Shichao Zhang, and Guan Yuan.
\newblock Bayesian graph local extrema convolution with long-tail strategy for misinformation detection.
\newblock {\em ACM Transactions on Knowledge Discovery from Data}, 18(4):1--21, 2024.

\bibitem[\protect\citeauthoryear{Zhang \bgroup \em et al.\egroup }{2024b}]{zhang2024survey}
Yi~Zhang, Yuying Zhao, Zhaoqing Li, Xueqi Cheng, Yu~Wang, Olivera Kotevska, S~Yu Philip, and Tyler Derr.
\newblock A survey on privacy in graph neural networks: Attacks, preservation, and applications.
\newblock {\em IEEE Transactions on Knowledge and Data Engineering}, 2024.

\bibitem[\protect\citeauthoryear{Zhang \bgroup \em et al.\egroup }{2025}]{zhang2025disentangled}
Guixian Zhang, Guan Yuan, Debo Cheng, Lin Liu, Jiuyong Li, and Shichao Zhang.
\newblock Disentangled contrastive learning for fair graph representations.
\newblock {\em Neural Networks}, 181:106781, 2025.

\end{thebibliography}
\appendix

\end{document}